%% file: acl2023.tex
\pdfoutput=1

\documentclass[11pt]{article}

\usepackage[review]{ACL2023}

\usepackage{amsmath}
\usepackage{amsfonts}
\usepackage{adjustbox}
\usepackage{booktabs}
\usepackage{bm}
\usepackage{graphicx}
\usepackage{latexsym}
\usepackage{makecell}
\usepackage{mathtools}
\usepackage{multirow}
\usepackage[switch]{lineno} 
\usepackage{pifont}
\usepackage{stackengine}
\usepackage{times}
\newcommand{\cmark}{\ding{51}}%
\newcommand{\xmark}{\ding{55}}%

\usepackage[T1]{fontenc}

\usepackage[utf8]{inputenc}

\usepackage{microtype}

\usepackage{inconsolata}

\title{Improving Cascaded Unsupervised Speech Translation with Denoising Back-translation}

\author{
Yu-Kuan Fu\textsuperscript{1}\Thanks{Equal Contribution},
Liang-Hsuan Tseng\textsuperscript{1}\footnotemark[1], 
Jiatong Shi\textsuperscript{2}, 
Chen-An Li\textsuperscript{1},
\\ 
{\bf Tsu-Yuan Hsu\textsuperscript{1}}, 
{\bf Shinji Watanabe\textsuperscript{2}}, 
{\bf Hung-yi Lee\textsuperscript{1}}
\\ 
\textsuperscript{1}College of Electrical Engineering and Computer Science, National Taiwan University \\
\textsuperscript{2}Language Technologies Institute, Carnegie Mellon University \\ 
\texttt{\textsuperscript{1}\{r11942083,r11921067,b08902123,b08201047,hungyilee\}@ntu.edu.tw} \\
\texttt{\textsuperscript{2}\{jiatongs@cs.cmu.edu,shinjiw@cmu.edu\}}
\\
}

\begin{document}
\nolinenumbers
{\makeatletter\acl@finalcopytrue
  \maketitle
}
\begin{abstract}
Most of the speech translation models heavily rely on parallel data, which is hard to collect especially for low-resource languages. To tackle this issue, we propose to build a cascaded speech translation system without leveraging any kind of paired data. We use fully unpaired data to train our unsupervised systems and evaluate our results on CoVoST 2 and CVSS. The results show that our work is comparable with some other early supervised methods in some language pairs. While cascaded systems always suffer from severe error propagation problems, we proposed denoising back-translation (DBT), a novel approach to building robust unsupervised neural machine translation (UNMT). DBT successfully increases the BLEU score by 0.7--0.9 in all three translation directions. Moreover, we simplified the pipeline of our cascaded system to reduce inference latency and conducted a comprehensive analysis of every part of our work. We also demonstrate our unsupervised speech translation results on the established website \footnote{\label{foot:us2st_demo} \href{https://anonymous-acl2023.github.io/us2s-demo/}{https://anonymous-acl2023.github.io/us2s-demo/}}.
\end{abstract}

\section{Introduction}
\input{section/introduction.tex}

\section{Related works}
\input{section/related_works.tex}

\section{Methods}
\input{section/methods.tex}

\section{Experiments}
\input{section/experiments.tex}
\section{Analysis}
\input{section/analysis.tex}

\section{Conclusion}

In this work, we build cascaded unsupervised speech-to-speech translation (US2ST) systems in several translation directions. To further improve the performance and mitigate the error propagation problems, we propose denoising back-translation (DBT), which is a novel method to improve the robustness of UNMT. DBT generally improves the performance of unsupervised speech translation (UST) across all the language pairs that we have experimented on. Without leveraging any paired data, our speech translation results are even better than some previous supervised methods. Additionally, we analyze the performance of each part in different settings individually; and we also attempt to integrate the TDN into the UNMT to reduce inference latency. In the future, we may investigate more techniques that can reduce the error propagation problems between different unsupervised cascaded modules; or conduct direct UST or US2ST.  


\section*{Limitations}
\input{section/limitation.tex}

\section*{Ethics Statement}
Our works build an effective UST cascaded system and try to mitigate the error propagation and inference latency. The communities might be interested in how to build a direct US2TT or even US2ST system or how to further improve the performance of the UST system.

\bibliography{anthology,custom}
\bibliographystyle{acl_natbib}




\end{document}

%% file: section/introduction.tex
Speech translation (ST) aims to convert speech from one language to another, allowing seamless communication between individuals speaking in different languages. Conventional speech-to-text translation (S2TT) system is accomplished by concatenating automatic speech recognition (ASR) and text-to-text machine translation (MT) \citep{ney1999speech} modules. Meanwhile, the cascaded speech-to-speech translation (S2ST) system further appends a text-to-speech (TTS) synthesis module after the S2TT system \citep{lavie1997janus, Wahlster2000VerbmobilFO,nakamura2006atr}. Recently, direct S2TT \citep{berard2016listen, weiss2017sequence} and S2ST \citep{jia2019direct, jia2021translatotron} systems emerge to solve the error propagation and inference latency problem of the cascaded systems. Some direct S2TT systems have shown comparable results or even outperform the cascaded S2TT systems \citep{wang2021large, bentivogli2021cascade}.

Most of the ST systems are trained on parallel data, which is extremely limited, especially for low-resource languages. This situation strongly hinders the performance of direct ST systems. Although cascaded systems could overcome this issue by collecting data for each component separately, they still face the challenge of domain mismatch caused by variations in data distribution across different corpora.

Compared to parallel data, unlabelled data is much easier to obtain regardless of modalities. The first unsupervised speech-to-text translation (US2TT) aligned spoken words with written words and then applied unsupervised word-by-word translation \citep{chung2018unsupervised, chung2019towards}. Moreover, with the recent progress in unsupervised automatic speech recognition (UASR) \citep{baevski2021unsupervised}, unsupervised neural machine translation (UNMT) \citep{lample2017unsupervised, lample2019cross, song2019mass}, and unsupervised text-to-speech (UTTS) synthesis \citep{ni2022unsupervised, liu2022simple}, \citet{wang2022simple} built an unsupervised speech-to-speech translation (US2ST) system. Besides building cascaded US2ST, they also generated pseudo labels for training direct US2TT systems. Their work might be considered concurrent with ours, mainly focusing on techniques to conduct simple and effective US2ST systems. 

Although the idea of cascaded US2ST is simple, directly concatenating UASR, UNMT, and UTTS might suffer from severe error propagation problems. For example, UNMT is trained on clean text, and small perturbations may greatly affect the translation results \citep{belinkov2017synthetic}. To tackle the issue, research on robust NMT has been widely investigated \citep{di2019robust, sperber2017toward, sun2020robust}; however, improving the robustness of UNMT is rarely studied. In this paper, we proposed denoising back-translation (DBT), a novel method to build a robust UNMT system. DBT combines the idea of denoising auto-encoding and back-translation (BT), dealing with error propagation issues in a fully unsupervised fashion. Briefly speaking, the pseudo text of DBT is generated from text with some noise, and the model should learn to reconstruct the clean text from the pseudo text. According to our results, this method substantially increases the quality of the cascaded unsupervised speech translation system.


Another issue with cascaded systems is the high inference latency. To address this, we integrated two parts of our cascaded system. First, the output of the UASR is normalized (stripped of punctuation marks and cases). Then, a text detokenizer is used to reconstruct the unnormalized text and feed it into the UNMT. By fine-tuning or continually training the UNMT with normalized source language text, the model is able to translate normalized source text into unnormalized target text. While there may be some degradation in performance, this method simplifies the pipeline of the cascaded system and significantly reduces inference time.

\input{figs/framework.tex}

We evaluate our cascaded US2ST system on CVSS \citep{jia-etal-2022-cvss}, a multilingual S2ST corpus; and CoVoST 2 \citep{wang2020covost}, a multilingual ST corpus of which CVSS built on top. We demonstrate that our US2ST could yield reasonable results across multiple translation directions, some of which are even better than the previous supervised approach\footnote{All data are public, and we will release the code, so the results will be easy to reproduce.}. Moreover, the proposed DBT method can help improve the performance by mitigating error propagation and domain mismatch problems.

%% file: figs/framework.tex
\begin{figure*}[ht!]
    \centering
    \includegraphics[width=0.78\linewidth]{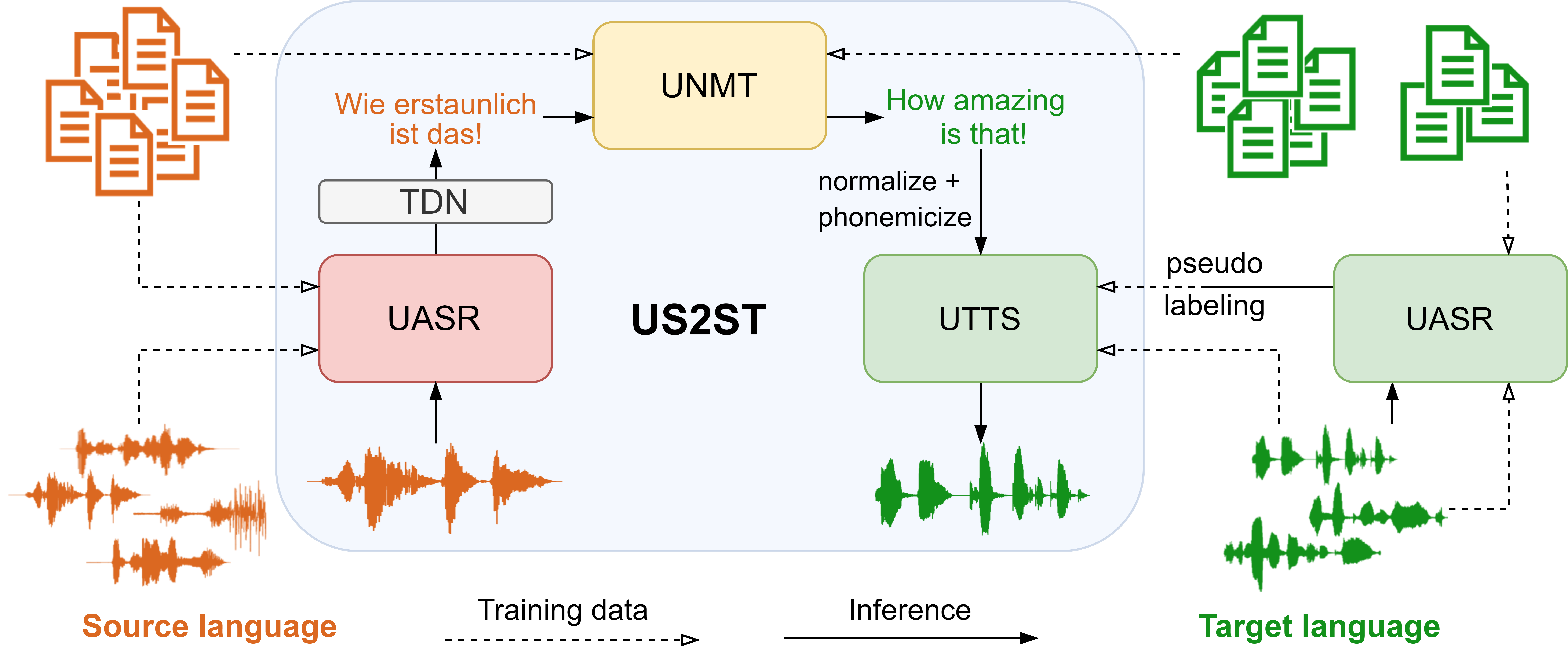}
    \caption{
        The framework of our cascade US2ST. 
    }
    \label{fig:framework}
\end{figure*}

%% file: section/related_works.tex
\subsection{ST}
Traditional S2TT system is composed of ASR and MT \citep{ney1999speech}, and S2ST system further append a TTS model after the MT model \citep{lavie1997janus, Wahlster2000VerbmobilFO,nakamura2006atr}. However, cascaded systems might suffer from error propagation and inference latency. Recent develpments in end-to-end S2TT \citep{berard2016listen, weiss2017sequence} and S2ST \citep{jia2019direct, jia2021translatotron} systems have been proposed to address these issues.

The main challenge of ST systems is the lack of parallel data. The unsupervised approach for ST is limited, but some progress has been made. The first US2TT system intended to learn the cross-modal alignment to map speech into written words, and used cross-lingual embedding to align words in different languages \citet{chung2018unsupervised, chung2019towards}. Additionally,  concurrent work built a US2ST system by combining SOTA UASR, UNMT, and UTTS to build a cascaded US2ST system. Further, they generated pseudo labels from the cascaded system to train an end-to-end US2ST system \citep{wang2022simple}.

\subsection{UASR}
UASR takes audio features or representations as input and generates phoneme sequences without supervision. To tackle the challenging problem, \citet{liu2018completely} first came out with the idea of applying a Generative Adversarial Network (GAN) \citep{goodfellow2020generative}. However, phoneme-level boundaries are required to segment the audio and construct embedding sequences. \cite{chen2019completely} breaks the limit by iteratively refining the audio segments with Hidden Markov Model (HMM) and GAN, achieving complete UASR. 

Recently, \citet{baevski2021unsupervised} proposed wav2vec-U, building the GAN-based UASR framework on top of the representation from wav2vec 2.0 (W2V2) \citep{baevski2020wav2vec}, a self-supervised speech model. The results outperformed previous SOTA, and are even comparable with some of the best-known supervised methods. Moreover, the original paper has shown that with the cross-lingual pre-trained version of W2V2 \citep{conneau2020unsupervised}, UASR in other languages is also available. The follow-up work, wav2vec-U 2.0 \citep{liu2022towards}, enabled the model to be trained end-to-end with the simplified pipeline and the improved training objective. 

\subsection{UNMT}
The first fully unsupervised neural machine translation model adopted a seq2seq model, and the encoder mapped monolingual corpus in two languages to a shared latent space via adversarial training. The decoder learned to reconstruct in both languages from the latent representations by denoising autoencoding loss and online back-translation loss \citep{lample2017unsupervised}. 

Recently, cross-lingual language model pretraining brought large progress to UNMT. XLM \citep{lample2019cross} first adopted masked language modeling pretraining to initialize the encoder and decoder, and then used back-translation loss together with denoising autoencoding loss to fine-tune the whole seq2seq model. MASS \citep{song2019mass} further used masked seq2seq pretraining to pretrain encoder and decoder jointly to reduce the discrepancy between pretraining and fine-tuning. Their proposed method can  align two languages with only back-translation loss. MASS outperforms XLM and all the other previous SOTA UNMT models in several language pairs.

\subsection{Robust NMT}
Traditional NMT and UNMT models are trained with clean input, thus small perturbations can greatly degrade the performance \citep{belinkov2017synthetic}. To improve the robustness, \citet{di2019robust, sperber2017toward} directly trained or fine-tuned the translation model on the target domain. 

As for the unsupervised scenario, \citet{sun2020robust} intended to improve the robustness of UNMT by applying some perturbation on positional embedding and word embedding to the input. The model learned to reconstruct the original input via denoising autoencoding loss and adversarial training.

\subsection{TTS and UTTS}
Some works intend to improve the performance of TTS through unlabeled data. For instance, pre-training the encoder/decoder \citep{chung2019semi}; utilizing the dual nature of TTS and ASR tasks \citep{ren2019almost}; applying variational auto-encoder to learn from speech disentanglement \citep{lian2022utts}. 

In spite of the improvement they brought, these methods still depend on certain levels of paired data. Directly training a UTTS without any supervision from paired data seems to be extremely hard. However, with recent success in UASR, UTTS might be accomplished in another way—training on the pseudo labels generated from UASR systems \citep{ni2022unsupervised, liu2022simple}.

%% file: section/methods.tex
Figure \ref{fig:framework} shows the architecture overview of our proposed approach to unsupervised speech-to-speech translation (US2ST). We split US2ST into three stages: unsupervised speech recognition (UASR), unsupervised machine translation (UNMT), and unsupervised speech synthesis (UTTS). The modules were trained separately but all in an unsupervised manner. During inference, we form the functionality of S2ST by concatenating them. Furthermore, we proposed denoising back-translation, to mitigate the error propagation and the domain mismatch issues between UASR and UMT submodules. 

\subsection{Base Architecture}

\paragraph{UASR}
\label{ssec:method_uasr}
We conducted our UASR subsystem following wav2vec-U \citep{baevski2021unsupervised}, and our code is based on their implementation in fairseq\footnote{\label{foot:fairseq} \href{https://github.com/facebookresearch/fairseq}{https://github.com/facebookresearch/fairseq}}. Besides its breakthrough performance on UASR in multiple languages, the robustness and stabilities across different corpora have also been well analyzed \citep{lin2022analyzing}. Thus, We mainly follow the data preparation procedure, model architecture, and the training objective of wav2vec-U. During inference, the model takes the preprocessed audio features as input and generates phoneme sequences. To further obtain word-level sequences, we adopt different decoding strategies, such as lexicon-based kenlm decoder and the weighted finite-state transducer (WFST; \citet{mohri2002weighted}). Self-training techniques on HMM are also applied for better performance. 

\paragraph{TDN} The outputs from the UASR are normalized word sequences. The UNMT model, however, may rely on punctuation marks and capital letters. For better performance, we also learned a text denormalizer to transform the generated word sequences back into sequences with punctuation marks and capitalized words, namely, denormalization. The module was a transformer-based seq2seq model. We first formed the paired data by normalizing raw text sentences and then trained the model with cross-entropy loss. 

\paragraph{UNMT}
UNMT aims to map sentences from source language $\mathcal{S}$ to target language $\mathcal{T}$ without leveraging any paired data. We conduct UNMT by following the architecture and pretrain process of MASS \citep{song2019mass}, which is a transformer-based seq2seq language model. During the fine-tuning process, we use denoising back-translation plus denoising autoencoder objective to align two languages and increase robustness.

\paragraph{UTTS}
We conducted UTTS by following the architecture of Variational Inference with adversarial learning for end-to-end Text-to-Speech (VITS), which has shown a significant performance gain over Tacotron2 and Transformer TTS in both subjective and objective evaluation \citep{kim2021conditional, hayashi2021espnet2}, and use the UASR system mentioned in section \ref{ssec:method_uasr} to generate pseudo labels for training.

\subsection{Mitigating Error Propagation}

\paragraph{Denoising back-translation.} \label{DBT} In our setting, the input of our translation model comes from the output of UASR, which might contain some noise, and worsen the performance significantly. In this paper, we introduce denoising back-translation, a novel approach for robust UNMT.

Given a source sentence $x \in \mathcal{S}$, a target sentence $y \in \mathcal{T}$, and $u^*(\cdot)$, $v^*(\cdot)$ are translation functions with directions of $\mathcal{S} \to \mathcal{T}$ and $\mathcal{T} \to \mathcal{S}$ respectively. we generate the pseudo parallel data by passing $x$ and $y$ through a noise function $f(\cdot)$, transcribed them into $u^*(f(x)) \in \mathcal{T}$ and $v^*(f(y)) \in \mathcal{S}$ respectively. $f(\cdot)$ can be some artificial data augmentations including deletion, insertion, or other modules like a language model. The objective of denoising back-translation is to reconstruct $x$ and $y$ from $u^*(f(x))$ and $v^*(f(y))$ respectively. The denoising back-translation loss is as follows:
\begin{equation}
\begin{split}
    \mathcal{L}_{DBT} = &\mathbb{E}_{x\in \mathcal{S}} [-\log{P_{\mathcal{T}\rightarrow \mathcal{S}}(x|u^{*}(f(x)))}] \\
     + &\mathbb{E}_{y\in \mathcal{T}} [-\log{P_{\mathcal{S}\rightarrow \mathcal{T}}(y|v^{*}(f(y)))}]
\end{split}
\end{equation}

Unlike the original back-translation, the pseudo labels of DBT are generated from noisy sentences, so the model should learn to transcribe the noisy pseudo sentences into clean sentences, and thus become more robust. The whole process of DBT was illustrated in Figure \ref{fig:dbt}.
\input{figs/dbt.tex}

%% file: figs/dbt.tex
\begin{figure}[t]
    \centering
    \includegraphics[width=0.8\linewidth]{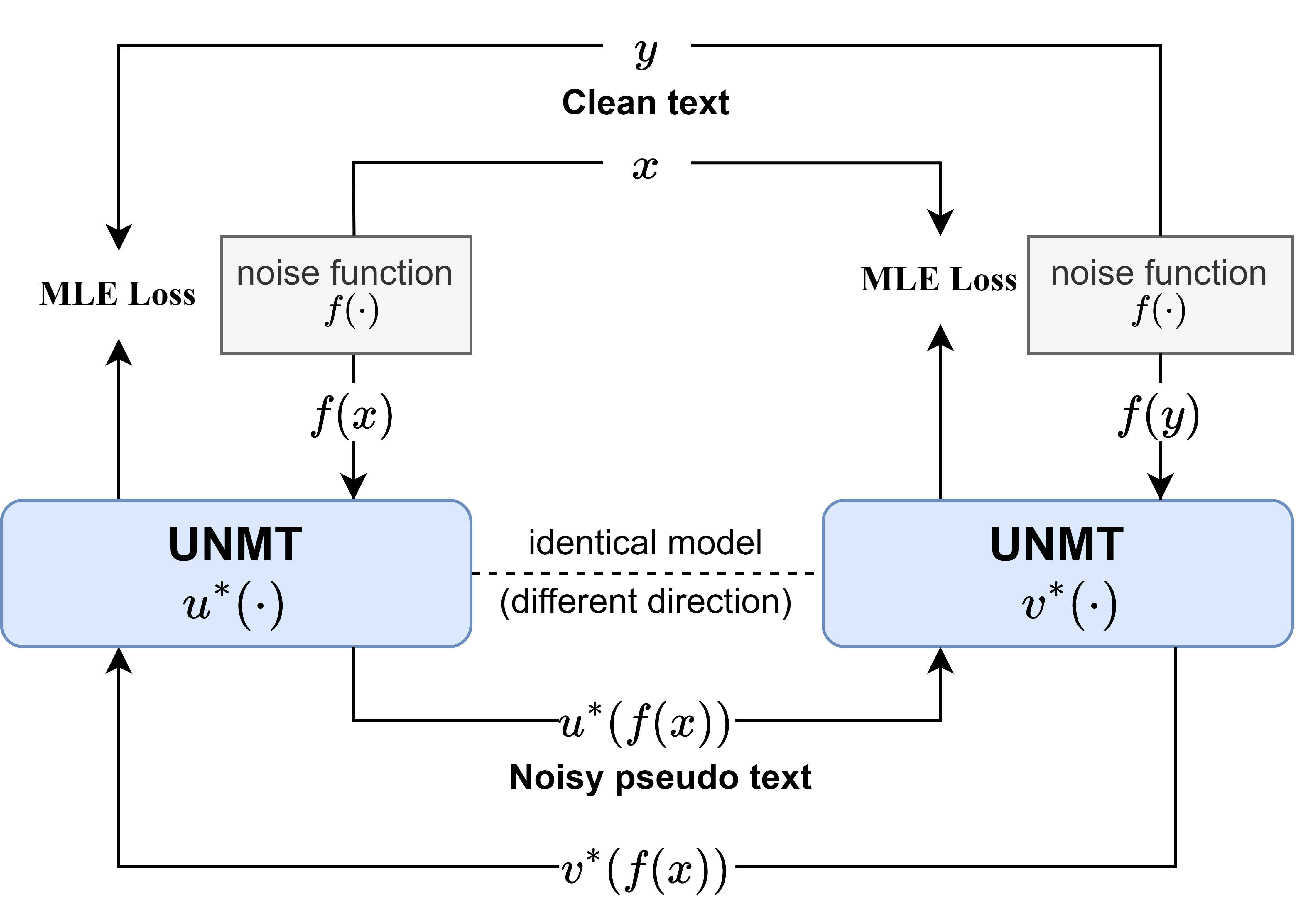}
    \caption{
        The illustration of our proposed method: DBT. 
    }
    \label{fig:dbt}
\end{figure}

%% file: section/experiments.tex
\subsection{Data}
To demonstrate our US2ST systems across different languages, we evaluate our S2ST results on CVSS, which is a multilingual corpus built on top of CoVoST 2 \citep{wang2020covost} and CommonVoice ver.4 (CV4; \citealp{ardila2019common}). Thus, we are also available to evaluate our S2TT results on CoVoST 2 and ASR results on CV4. 

However, \textbf{we do not utilize any paired data from the corpus during training}; instead, we use audio and text data from different corpora, constructing an unpaired scenario for our US2ST. For audio, we adopt Common Voice ver.4 for UASR and LJspeech \citep{ljspeech17} for UTTS without using any transcriptions from them; and for text, we extract sentences from Wikipedia\footnote{\label{foot:wiki}We  extract the data from wiki using \href{https://github.com/attardi/wikiextractor}{WikiExtractor} \citep{Wikiextractor2015}}, WMT'14, CC100 \citep{conneau2019unsupervised}, and LibriSpeech LM data\footnote{\label{foot:LS}Following \citet{liu2022simple}, we exclude the transcriptions of LJspeech to form fully unpaired scenario}. All of the data we used is open-sourced and public-available. 

\subsection{System setups}

\paragraph{UASR}
We used audio from CV4 and text data from Wikipedia to train our UASR models. More precisely, we use 100 hours of audio and about 1--3M sentences for each language. After training, we evaluate the results with the transcriptions from CV4. For the pre-trained W2V2 model, we directly use the cross-lingual version (XLSR; \citealp{conneau2020unsupervised}) without finetuning. XLSR had pre-trained in many different languages thus suiting our needs for training UASR in languages other than English. 

During preprocessing, we adopted the same configuration in wav2vec-U, except for the silence insertion rate of French. We found that our French model converged better when <SIL> token insertion rate is 0.5 instead.

As for the GAN training configuration, we chose the coefficients of the loss function according to the original paper as follows: the gradient penalty weight $\lambda=1.5$ or $2.0$, the smoothness penalty weight $\gamma=0.5$, and the phoneme diversity loss weight $\eta=4$. We trained $3$ seeds for each configuration, conducting 6 models for each language. 

\paragraph{TDN} For the text denormalizer (TDN), we adopted transformer encoder-decoder architecture, both of which contained 4 layers of transformer blocks. We constructed the input data by normalizing 8--10M of plain text data from CC100, and the objective is to reconstruct the unnormalized data. 

\paragraph{UNMT}
We used the back-translation fine-tuned MASS model released by Microsoft\footnote{\label{foot:MASS}\href{https://github.com/microsoft/MASS}{https://github.com/microsoft/MASS}} to initialize our German-English and French-English UNMT models. For the Spanish-English UNMT model, we followed the same pretraining and fine-tuning steps as the standard MASS model.

For denoising back-translation as discussed in Section \ref{DBT}, the artificial noise $f(\cdot)$ included random drop, substitution, and insert, whose probability were 0.05, 0.01, and 0.05 respectively. We continually fine-tune the model by denoising back-translation loss plus denoising autoencoding loss to build a robust UNMT model.

\paragraph{UTTS}
In this paper, we conducted single-speaker UTTS, so we trained a UASR model on the audio of LJSpeech \citep{ljspeech17}. Using the UASR-generated pseudo labels, we then proceeded with the training of VITS\footnote{\label{foot:VITS}\href{https://github.com/jaywalnut310/vits}{https://github.com/jaywalnut310/vits}}, but the transcription of LJSpeech was replaced by the pseudo label.

\input{tabs/S2ST.tex}
\subsection{Supervised cascaded S2ST}
First, we constructed our upper bound model by training a supervised cascaded S2ST (ASR$\to$MT$\to$TTS) which shares similar model architecture with our US2ST. 

For ASR, we finetuned the whole XLSR instead of treating it as a feature extractor. We adopted letter-based training and followed the configuration from fairseq  \citep{ott2019fairseq}. The amount of audio data was exactly the same as those in UASR. Furthermore, we finetuned the XLSR models individually for each language. MT is achieved by training the same initial model with UNMT, but the training data were the transcriptions of CoVoST 2 plus CC100. We supervised trained with CoVoST 2, and CC100 was used for back-translation training, which can boost the performance of supervised machine translation models. Finally, instead of training on pseudo labels from UASR, the supervised TTS model directly uses the reference phoneme sequences and their corresponding utterance audio as paired data. 

By constructing cascaded supervised S2ST, we can discuss the performance individually for each component.

\subsection{Evaluation} The evaluation metric of ASR was normalized word error rate (WER), which removed all punctuation marks and converted all characters to lowercase. We used sacreBLEU\footnote{\href{https://github.com/mjpost/sacrebleu}{https://github.com/mjpost/sacrebleu}} to calculate the BLEU score of S2TT. For the final S2ST results, Whisper
\footnote{we used the cross-lingual large-v2 model released by \href{https://github.com/openai/whisper}{https://github.com/openai/whisper}}
\citep{radford2022robust}, a supervised ASR model released by OpenAI, is adopted to transcribe the hypothesized audio and calculated the BLEU score.

\subsection{Results}
We show our results in Table \ref{table:all}, including the results of ASR, S2TT, and S2ST. To compare our US2ST system with others' works, we collect some results from the previous studies on CoVoST 2 and CVSS (\textit{(a)}--\textit{(f)}, \textit{(h)}--\textit{(i)}). To get better comparisons, both cascaded and direct systems are included.

Next, we discuss the details of the methods in the table. \textit{(a)} comes from \citet{wang2020covost}; among all experiments in their paper, we only report the results of the cascaded S2TT system constructed by monolingual ASR and bilingual MT for fair comparison. According to the table, our US2TT performances in De--En are just having small degradation from theirs (\textit{(j)} and \textit{(k)} vs \textit{(a)}), indicating that our US2TT system can be comparable to some early supervised cascaded S2TT works in some language pairs.

Rows \textit{(b)} and \textit{(c)} are the results of the direct S2TT systems from \citet{wang2020fairseq}. They have developed a tool-kit for S2TT and demonstrated it on CoVoST 2 with different model backbones. We compare our US2TT results (row \textit{(j) and \textit{(k)}}) with their Transformer-based models. Our results have not only outperformed their small model (\textit{(b)}) in De-En and Es-En tasks but also outperformed their large model in De-En (row \textit{(c)}). Our works are comparable to some early direct S2TT systems (\textit{(b)}, \textit{(c)}) except for the Fr-En pair.

Row \textit{(d)} is the SOTA direct S2TT model, which is a cross-lingual speech model based on wav2vec 2.0 architecture. Taking advantage of the large pretrained self-supervised model, the performance is significantly better than all the other works in all language pairs, showing there is still a huge gap between SOTA supervised S2TT and our US2TT systems.

Rows \textit{(h)} and \textit{(i)} come from the concurrent US2ST system \citep{wang2022simple}; \textit{(h)} is constructed by concatenating UASR, UNMT, and UTTS, and \textit{(i)} is an end2end S2ST systems trained on pseudo labels generated by \textit{(h)}. Our model architecture is similar to theirs, while they fine-tune the wav2vec 2.0 with target languages in UASR to mitigate the domain mismatch between pretraining and downstream tasks. For Fr-En S2TT, their performance is superior to ours. This may be due to the difficulty of French UASR, which has a higher error rate than other languages. This phenomenon was also observed in \citet{wang2022simple}, and we think that it might be more severe in our case since we did not fine-tune the wav2vec 2.0 models on the target languages. Suffering from domain mismatch, our systems still outperform their works in Es-En S2TT. 

The performance of our S2ST systems drops even more than S2TT. However, this is because we did not utilize the data from CVSS when training our UTTS models for the fairness concern. Other studies (\textit{(e)}, \textit{(f)}, \textit{(h)}, \textit{(i)}) directly trained the whole model or UTTS with the data from CVSS. Since our models are trained with LJSpeech (row \textit{(g)}, \textit{(j)}, \textit{(k)}), they could have severe domain mismatch problems during inference on out-domain data. We had also investigated the cause of the performance drop in the later section and came out with the same conclusion. In spite of this, our unsupervised S2ST results still outperform Translatotron (row \textit{(j)}, \textit{(k)} v.s. \textit{(e)}) in De-En and Es-EN translation directions.

Last but not least, Our cascaded system with DBT \textit{(k)} outperforms \textit{(j)} by 0.7 to 0.9 in both US2TT and US2ST, indicating a model can better translate the noisy input under the guidance of DBT, and thus mitigate the problem of error propagation in cascaded systems.

%% file: tabs/S2ST.tex
\begin{table*}[hbt!]
    \caption{The results are evaluated on CoVoST 2 for S2TT, and CVSS for S2ST. C-T stands for cascaded S2TT; D-S for direct S2ST.}
    \centering
    \vspace{1pt}
    \begin{adjustbox}{width=\columnwidth*2,center}
    \begin{tabular} {l c|@{\extracolsep{4pt}}ccc ccc ccc@{}}
        \toprule
        \multirow{2}{*}{Method} & \multirow{2}{*}{Type} & \multicolumn{3}{c}{ASR $\downarrow$} &
        \multicolumn{3}{c}{S2TT (X$\to$En) $\uparrow$} & \multicolumn{3}{c}{S2ST (X$\to$En) $\uparrow$} \\
        \cline{3-5}
        \cline{6-8}
        \cline{9-11}
          &  & Fr & De & Es 
        & Fr & De & Es 
        & Fr & De & Es \\
        \midrule
        \multicolumn{10}{l}{\textbf{\large{\textsc{Supervised learning}}}} \\
        \textit{(a)} \citet{wang2020covost} & C-T
        & 18.3 & 21.4 & 16.0
        & 27.6 & 21.0 & 27.4
        & - & - & - \\
        \textit{(b)} fairseq S2T (T-Sm) \citep{wang2020fairseq} & D-T 
        & - & - & - 
        & 26.3 & 17.1 & 23.0
        & - & - & - \\
        \textit{(c)} fairseq S2T (Multi. T-Md) & D-T
        & - & - & - 
        & 26.5 & 17.5 & 27.0
        & - & - & - \\
        \textit{(d)} XLS-R (2B) \citep{babu2021xls} & D-T
        & - & - & - 
        & 37.6 & 33.6 & 39.2 
        & - & - & - \\
        \textit{(e)} Translatotron \citep{jia2019direct}  & D-S
        & - & - & - 
        & - & - & - 
        & 15.5 & 6.9 & 14.1 \\
        \textit{(f)} Translatotron 2 \citep{jia2021translatotron}  & D-S
        & - & - & - 
        & - & - & - 
        & 28.3 & 19.7 & 23.5 \\
        
        \textit{(g)} Our upper bound & C-S
        & 16.2 & 14.1 & 11.0 
        & 29.5 & 25.6 & 31.0
        & 23.8 & 21.8 & 26.2 \\
        \midrule
        \midrule
        \multicolumn{10}{l}{\textbf{\large{\textsc{Unsupervised learning}}}} \\
        \textit{(h)} \citet{wang2022simple} cascaded US2ST &  
        & - & - & - 
        & 24.4 & - & 23.4 
        & 21.6 & - & 21.2 \\
        \textit{(i)} \citet{wang2022simple} end-to-end US2ST &  
        & - & - & - 
        & 24.2 & - & 24.0 
        & 21.2 & - & 20.1 \\
        \textit{(j)} Our cascaded US2ST  & C-S
        & \multirow{2}{*}{33.2} & \multirow{2}{*}{23.8} & \multirow{2}{*}{17.4}
        & 20.0 & 19.5 & 23.8
        & 13.4 & 13.8 & 16.7 \\
        \textit{(k)} Our cascaded US2ST with DBT (artificial noise) & C-S
        &  &  & 
        & 20.8 & 20.4 & 24.5
        & 14.4 & 14.7 & 17.4 \\
        \bottomrule
    \end{tabular}
    \end{adjustbox}
    \label{table:all}
\end{table*}

%% file: section/analysis.tex
\label{sec:analysis}
\paragraph{Stabilities of UASR cross different languages} First of all, we found that the stabilities of our UASR models vary between languages. The measurement of the stability is by calculating the percentage of the converged rate among the models leveraging the same amount of text and speech data. We consider a UASR model is converged if its $PER<50\%$. We summarize the discoveries in Table \ref{tab:uasr_stability}. According to our experiments, German and Spanish are easier to converge; while French usually can not converge well. However, we found that it might be more suitable for French UASR models to converge if we change the <SIL> token insertion rate from $0.25$ to $0.5$. 
\input{tabs/UASR/stability}

\paragraph{Decoding and self-training in UASR} The original outputs of wav2vec-U are in phoneme-level, which are incompatible with the UNMT. However, with the integration with LM, we are available to obtain word-level output sequences. As shown in the part \textbf{(I)} of Table \ref{tab:uasr_decode}, we demonstrate that the two decoding methods, Kenlm and WFST can both generate word sequences. The second part \textbf{(II)} in the table illustrates the effectiveness of self-training on HMM. Among all the methods, we considered that the best strategy we found was by conducting self-training on HMM with the pseudo labels from WFST decoding. More surprisingly, even if the pseudo labels come from Viterbi decoding, using these labels on HMM can make huge improvements. After self-training, the performance gap between Viterbi and WFST decoding became relatively small. Note that for simplicity, we only show the results on the testing set of CV4-German; while the results on other languages also share similar trends.
\input{tabs/UASR/decode}

\paragraph{Integration of text denormalization and UNMT}
In this section, we try to integrate text denormalization into UNMT; reducing pipelines of cascaded system might mitigate error propagation, and reduce inference time. 

We introduced normalized fine-tuning (NFT) to direct translate normalized source text into unnormalized target text. NFT initializes the model as original DBT fine-tuning, but fine-tunes on normalized source text and unnormalized target text. While directly fine-tune the checkpoint pretrained on unnormalized text might induce mismatch between pretraining and fine-tuning. To address this problem, we further introduce normalized continual training (NCT), which continual pretrains the checkpoint on normalized source text and unnormalized target text, and follow NFT for downstream task fine-tuning. 

The results are shown in Table \ref{tab:udn}. we compare NFT (I) and NCT (II) with our two baselines: (I) UDN + UNMT (our original setting), (II) UNMT (only use the translation model of (I)). The performance of (II) drops a lot, for normalized text never appear during the training process of the model, directing translating on that induces severe domain mismatch. (III) and (IV) have outperformed (II) a lot, but they still decrease the BLEU score by about 2.6 and respectively ((III), (IV) v.s. (I)), indicating mismatch between pretraining and downstream task training is severe, and NCT has only minor improvement.

Integrating text denormalization and UMT, while it did not performs better due to the mismatch between pretraining and fine-tuning, can still reduce the inference latency. To address this mismatch, a potential solution would be pretraining the model on normalized source text and unnormalized target text from scratch.

\paragraph{Robustness of UNMT}
\input{tabs/RUNMT.tex}
DBT has been shown to improve the performance of cascaded S2TT and S2ST systems. In this section, we investigate the robustness of UNMT. Table \ref{tab:rumnt} shows the BLEU scores of translating the ground truth of CoVoST ("Clean"), and that of translating the output of UASR ("ASR"). 

\input{tabs/integration.tex}

The results indicate that for Fr-En and Es-En on "Clean", the performance drops slightly compared to that of BT, but the score drop from "Clean" to "ASR" of DBT decreased by about 1 BLEU score. For De-En, DBT even performs better than BT on "Clean", and the performance drops from "Clean" to "ASR" are the same, which means that DBT can be regraded as a new data-augmentation method to boost the performance of a UNMT model. This result from avoiding the model from directly copying the input during generating pseudo-label for back-translation.

Overall, DBT increases the robustness without sacrificing its performance on clean input too much, and it even outperforms BT on "Clean" in some cases.

\input{tabs/utts.tex}

\paragraph{Performance analysis of UTTS}
In this section, we present more analytical results of our UTTS submodule. In part \textit{(I)} in Table \ref{tab:utts}, we evaluate our supervised TTS and UTTS on in-domain testing set (LJspeech) and out-domain testing set (US2ST, Fr$\to$En). After we got the synthesis speech, we send the audio to whisper and then calculate the WER. We used the base model for this experiment to further accentuate the performance differences. Our results indicated that the performance drop between supervised TTS and UTTS is much lower than the error induced by the domain mismatch problem. The results also emphasized that the domain mismatch between the training data of TTS models and the testing data is one of the main reasons for our S2ST performance drop. Leveraging the data from CVSS for UTTS training might be a solution, but it may also induce fairness concerns from our point of view.  

Next, in section \textit{(II)}, we evaluate our supervised TTS and UTTS models on the outputs from supervised ST and UST. According to the table, we can infer that the gap between supervised TTS and UTTS might be overestimated. The performance drop induced by pseudo-labeling is acceptable or at least reasonable. 

%% file: tabs/UASR/stability.tex
\begin{table}[h!]
    \caption{Stabilities of UASR across different languages. }
    \centering
    \small
    \begin{adjustbox}{width=\columnwidth,center}
    \begin{tabular}{l|ccc}
        \toprule
        Lang. & \makecell{<SIL> \\ ins. rate}  & \makecell{Best PER \\ (\textit{Viterbi})} & \makecell{\%-converged \\ ($\text{PER}<50\%$)} \\
        \midrule
        De & 0.25 & 25.3\% & 66\% \\
        \midrule
        Es & 0.25 & 27.0\% & 50\% \\
        \midrule
        Fr & 0.25 & 49.2\% & <10\% \\
         & 0.50  & 35.2\% & 17\% \\
        \bottomrule 
    \end{tabular}
    \end{adjustbox}
    \label{tab:uasr_stability}
\end{table}

%% file: tabs/UASR/decode.tex
\begin{table}[t]
    \centering
    \caption{Comparison of different decoding strategies and the improvement brought by HMM self-training. We use the same 4-gram LM (phoneme-level or word-level) across different methods. }
    \begin{tabular}{lccc}
        \toprule
        Method & LM & PER(\%) & WER(\%) \\
        \midrule
        \multicolumn{3}{l}{\textbf{(I) }\textit{\textbf{Without self-training}}} \\
        Viterbi & \xmark & 25.2 & - \\
        Kenlm & \cmark &29.5 & 39.5 \\
        WFST & \cmark & 21.3 & 34.4 \\
        \midrule
        \midrule
        \multicolumn{3}{l}{ \textbf{(II) }\textit{\textbf{With self-training}}} \\
        Viterbi$\to$ HMM & \cmark & 15.2 & 25.3 \\
        WFST$\to$ HMM & \cmark & \textbf{14.4} & \textbf{23.8} \\
        \bottomrule 
    \end{tabular}
    \label{tab:uasr_decode}
\end{table}

%% file: tabs/RUNMT.tex
\begin{table}[h!]
\caption{Robustness of UNMT across languages. "Clean" and "ASR" refer to the BLEU score of translating on the ground truth and UASR output respectively.}
\centering
\begin{tabular}{c|@{\extracolsep{4pt}}cccc@{}}
\toprule
\multirow{2}{*}{Direction} & \multicolumn{2}{c}{Clean} & \multicolumn{2}{c}{ASR} \\ \cline{2-3} \cline{4-5} 
& BT          & DBT         & BT            & DBT          \\ 
                           \midrule
Fr-En        & 35.3        & 35.0        & 20.0          & 20.8         \\
De-En        & 27.1        & 28.0        & 19.5           & 20.4          \\
Es-En        & 33.4        & 33.1        & 23.8           & 24.5 \\     
\bottomrule
\end{tabular}
\label{tab:rumnt}
\end{table}

%% file: tabs/integration.tex
\begin{table}[]
    \centering
    \caption{The BLEU score of integrating text normalization and UNMT for De-En S2TT. }
    \begin{tabular}{lc}
    \toprule
    \multicolumn{1}{c}{Model} & S2TT \\
    \midrule
    (I) TDN + UNMT & 20.8         \\
    (II) UNMT  & 13.0         \\
    \midrule
    (III) NFT  & 18.2         \\
    (IV) NCT   & 18.6         \\
    \bottomrule
    \end{tabular}

    \label{tab:udn}
\end{table}

%% file: tabs/utts.tex
\begin{table}[h]
    \caption{Analysis of our UTTS models. In part \textit{(I)}, we the performance drop due to the domain mismatch problem. In part \textit{(II)}, we further investigate the effectiveness of our UTTS by using the same testing data as the supervised TTS. }
    \centering
    \begin{adjustbox}{width=\columnwidth,center} \renewcommand{\arraystretch}{1.15}
    \begin{tabular}{l |cc}
        \toprule
        Testing data & TTS & UTTS \\
        \midrule
        \midrule
        \multicolumn{3}{l}{\textit{(I) WER on in-domain / out-domain data.}} \\
        \midrule
        in-domain  & 23.5\% & 31.5\% \\
        out-domain & 46.5\% & 54.2\% \\
        \midrule
        \midrule
        \multicolumn{3}{l}{\textit{(II) BLEU score of using ST / UST (Fr$\to$En)}} \\
        \midrule
        sup. ST   & 23.8 & 20.6  \\
        unsup. ST (BT) & 15.5 & 13.4  \\
        \bottomrule
    \end{tabular}
    \end{adjustbox}
    \label{tab:utts}
\end{table}

%% file: section/limitation.tex
%
In this work, we have handled the problem of error propagation among UASR, TDN, and UNMT. Nevertheless, we didn't resolve that between UTTS and other modules, which may lead to a lower score of the S2ST result.

Our methodology works for most languages, however, our US2ST is based on UNMT for unpaired text data. Therefore, it is limited to written languages. We believe that our denoise back-translation brings new insights to US2ST and can extend to unwritten language setups.